\documentclass[letterpaper]{article} % DO NOT CHANGE THIS
\usepackage{main}  % DO NOT CHANGE THIS
\usepackage{times}  % DO NOT CHANGE THIS
\usepackage{helvet}  % DO NOT CHANGE THIS
\usepackage{courier}  % DO NOT CHANGE THIS
\usepackage[hyphens]{url}  % DO NOT CHANGE THIS
\usepackage{graphicx} % DO NOT CHANGE THIS
\usepackage{amsmath}
\usepackage{booktabs}
\usepackage{multirow}
\usepackage{arydshln} % 支持虚线 \cdashline
\usepackage{array}
\usepackage{amssymb}
\usepackage{subcaption}
\usepackage{natbib}  % DO NOT CHANGE THIS AND DO NOT ADD ANY OPTIONS TO IT
\usepackage{caption} % DO NOT CHANGE THIS AND DO NOT ADD ANY OPTIONS TO IT
\usepackage{algorithm}
\usepackage{algorithmic}

\usepackage{newfloat}
\usepackage{listings}
\DeclareCaptionStyle{ruled}{labelfont=normalfont,labelsep=colon,strut=off} % DO NOT CHANGE THIS
\floatstyle{ruled}
\newfloat{listing}{tb}{lst}{}
\floatname{listing}{Listing}
\title{Semantic-Consistent Bidirectional Contrastive Hashing for Noisy Multi-Label Cross-Modal Retrieval}
\author{
    Likang Peng\textsuperscript{\rm 1},
    Chao Su\textsuperscript{\rm 1},
    Wenyuan Wu\textsuperscript{\rm 1},
    Yuan Sun\textsuperscript{\rm 2},
    Dezhong Peng\textsuperscript{\rm 1,3},
    Xi Peng\textsuperscript{\rm 1,3},
    Xu Wang\textsuperscript{\rm 1,4}\thanks{Corresponding author}
}
\affiliations{
    \textsuperscript{\rm 1}The College of Computer Science, Sichuan University, Chengdu, China\\
    \textsuperscript{\rm 2}National Key Laboratory of Fundamental Algorithms and Models \\ for Engineering Numerical Simulation, Sichuan University, Chengdu, China\\
    \textsuperscript{\rm 3}Tianfu Jincheng Laboratory, Chengdu, China\\
    \textsuperscript{\rm 4}Centre for Frontier AI Research (CFAR), A*STAR, Singapore\\
    likangpeng@stu.scu.edu.cn, 
    suchao.ml@gmail.com,
    wuwenyuan97@gmail.com,
    sunyuan\_work@163.com,
    pengdz@scu.edu.cn,
    pengx.gm@gmail.com,
    wangxu.scu@gmail.com
}

\begin{document}

\maketitle

\begin{abstract}
Cross-modal hashing (CMH) facilitates efficient retrieval across different modalities (e.g., image and text) by encoding data into compact binary representations. While recent methods have achieved remarkable performance, they often rely heavily on fully annotated datasets, which are costly and labor-intensive to obtain. In real-world scenarios, particularly in multi-label datasets, label noise is prevalent and severely degrades retrieval performance. Moreover, existing CMH approaches typically overlook the partial semantic overlaps inherent in multi-label data, limiting their robustness and generalization. To tackle these challenges, we propose a novel framework named Semantic-Consistent Bidirectional Contrastive Hashing (SCBCH). The framework comprises two complementary modules: (1) Cross-modal Semantic-Consistent Classification (CSCC), which leverages cross-modal semantic consistency to estimate sample reliability and reduce the impact of noisy labels; (2) Bidirectional Soft Contrastive Hashing (BSCH), which dynamically generates soft contrastive sample pairs based on multi-label semantic overlap, enabling adaptive contrastive learning between semantically similar and dissimilar samples across modalities. Extensive experiments on four widely-used cross-modal retrieval benchmarks validate the effectiveness and robustness of our method, consistently outperforming state-of-the-art approaches under noisy multi-label conditions.
\end{abstract}

\begin{links}
    \link{Code}{https://github.com/Plke/SCBCH}
    % \link{Datasets}{https://aaai.org/example/datasets}
    % \link{Extended version}{https://aaai.org/example/extended-version}
\end{links}

\section{Introduction}

The landscape of digital data has expanded from traditional text to multimodal content like images and videos, driven by internet advancements that complicate information retrieval. Cross-modal retrieval \cite{DiCA,DiDA} tackles this challenge by enabling searches across diverse modalities. For scalable and efficient large-scale retrieval across modalities, cross-modal hashing \cite{DCMH_super_2017,mlhashing} transforms heterogeneous data into a shared Hamming space through compact binary codes, allowing fast search with low computational overhead.

However, most existing methods rely on clean and fully annotated training data \cite{multi_modal_survey,RoDA,wu2025improving}. Such data are expensive to obtain in the real world, due to inherent data complexity \cite{data_complexity} and human annotation errors \cite{human_annotation_errors}. While unsupervised methods reduce labeling costs, they often underperform compared to supervised approaches.
A practical compromise is to use non-expert annotations, which inevitably introduce label noise, particularly in cross-modal multi-label scenarios. 

\begin{figure}[t]
    \centering
    \includegraphics[width=0.9\linewidth]{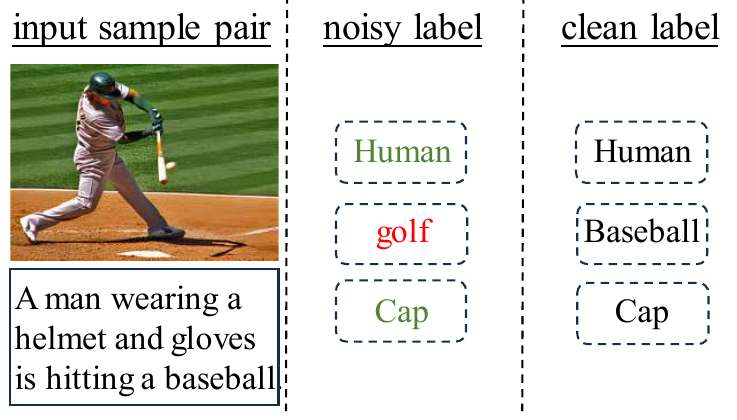} 
    \caption{Cross-modal (image-text) pair with noisy multi-label annotations, where green indicates correct labels and red indicates incorrect ones.}
    \label{fig:introduction}
\end{figure}

As illustrated in Figure \ref{fig:introduction}, an image-text pair is annotated with three labels, one of which is noisy: the label ``golf" is incorrect due to an annotation error. While robust learning strategies such as sample selection \cite{human_annotation_errors} and loss correction \cite{Symmetric_robust_loss_2019} have been explored in single-modal tasks, they are not directly applicable to cross-modal settings due to modality heterogeneity. 
To address label noise in cross-modal retrieval while maintaining scalability and retrieval efficiency, several noise-robust hash learning methods \cite{sup_DHRL_TBD2024,MetaVG,Graph} have been proposed. Although these methods aim to improve robustness, they often introduce new trade-offs. For example, \cite{sup_DHRL_TBD2024} mitigates noise by excluding suspected noisy samples from training, which may inadvertently discard valuable supervisory signals.
Meanwhile, \cite{NRCH} treats any pair with at least one shared label as a positive pair, and all others as negative. This binary treatment may overlook the nuanced semantic relationships common in multi-label scenarios, potentially resulting in imbalanced pair construction and underutilization of ambiguous yet informative samples.

To overcome these limitations, we propose a novel framework called Semantic-Consistent Bidirectional Contrastive Hashing (SCBCH), which jointly leverages Cross-modal Semantic-Consistent Classification (CSCC) and Bidirectional Soft Contrastive Hashing  (BSCH) objective to improve robustness under noisy supervision.
Specifically, in CSCC, for each anchor sample from both modalities, we select its most similar neighbors in the feature space. The label similarities with these neighbors serve as soft weights to compute a weighted classification loss, enabling more reliable supervision under noisy labels.
Meanwhile, BSCH leverages partial label overlap in multi-label settings to construct soft contrastive pairs, instead of relying on hard positives or hard negatives. This allows for a more nuanced assessment of semantic similarity between samples. The resulting dynamic pairing strategy enables the formation of more reliable sample pairs, effectively mitigating the impact of noisy annotations and enhancing the model’s robustness.
The main contributions of the proposed SCBCH framework are summarized as follows:

% framework
\begin{figure*}[ht]
\centerline{\includegraphics[width=0.9\linewidth]{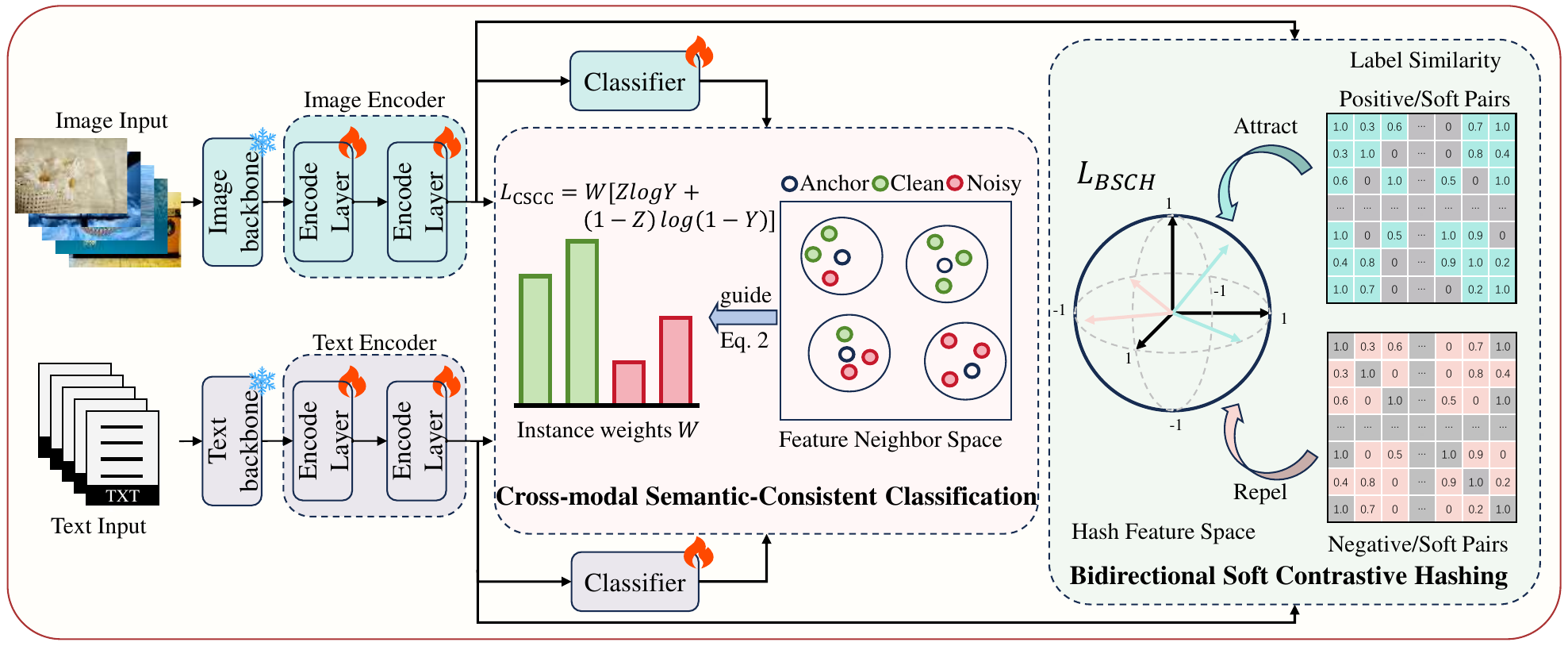}}
    \caption{The SCBCH framework for cross-modal hashing under noisy multi-label supervision consists of CSCC and BSCH. CSCC enhances label reliability by leveraging cross-modal neighbor consistency to adaptively reweight samples, while BSCH employs bidirectional contrastive learning to construct reliable soft pairs based on multi-label overlap, explicitly attracting similar pairs and repelling dissimilar ones to improve robustness against noise.
    }
    \label{fig:framework}
\end{figure*}

\begin{itemize}
\item We study an underexplored problem of learning cross-modal hash codes under noisy multi-label supervision, and propose Semantic-Consistent Bidirectional Contrastive Hashing (SCBCH), a framework designed to improve retrieval performance under label noise.

\item A Cross-modal Semantic-Consistent Classification (CSCC) module is introduced to exploit semantic consensus from neighbors, adaptively reweighting all samples instead of filtering, thereby improving training stability and robustness in the presence of label noise.

\item We propose a novel Bidirectional Soft Contrastive Hashing (BSCH), a contrastive learning paradigm tailored for multi-label scenarios that dynamically constructs soft pairs based on label overlap to capture fine-grained semantics and enhance robustness.

\item Extensive experiments conducted on four widely used benchmark datasets demonstrate the effectiveness of our approach. Compared with 12 state-of-the-art baselines, SCBCH achieves consistently superior robustness under various levels of label noise.
\end{itemize}

\section{Related Work}

\subsection{Noisy Label Learning}
To address the challenge of learning with noisy labels, a variety of strategies have been proposed, which can be broadly categorized into two main approaches: robust learning and clean sample selection. For robust learning~\cite{wang2025noisy}, since the conventional cross-entropy loss is sensitive to label noise, several studies have introduced noise-tolerant loss functions \cite{Generalized_robust_loss_2018} and regularization techniques \cite{regularization_1_2020,regularization_2_2020} to enhance model robustness.
For example, the symmetric cross-entropy loss \cite{Symmetric_robust_loss_2019} incorporates a noise-resistant objective that facilitates faster convergence and improved performance. However, empirical studies have shown that robust learning methods often face a trade-off between noise tolerance and predictive accuracy \cite{balance_noisy_2022}.
In contrast, sample selection methods aim to identify clean instances from noisy datasets based on predefined criteria. 
For instance, MentorNet \cite{MentorNet_2019} introduces an auxiliary network to filter out noisy samples, while DivideMix \cite{Dividemix_2020} employs a Gaussian Mixture Model to dynamically divide training data into clean and noisy subsets. 
Nevertheless, these methods inevitably overlook hard examples with high loss that may still be informative, and they are less effective at bridging the modality gap in cross-modal tasks, as they are primarily designed for single-modality settings.

\subsection{Cross-modal Hashing}
Numerous cross-modal hashing methods have emerged, mainly categorized into unsupervised and supervised approaches. Unsupervised methods \cite{PIP,unsup_CIRH_zhu2022work} learn a shared Hamming space by leveraging data structure without labels. For example, UCCH \cite{hu2022UCCH} incorporates contrastive learning with a triplet ranking loss to reduce modality gaps. However, the absence of supervision often limits their performance.
In contrast, supervised methods \cite{wang2020deep,wang2021mars,su2025neighboraware} assume that all training labels are clean and accurate, and generally optimize classification or similarity-preserving objectives, such as pairwise losses \cite{DCMH_super_2017} and triplet losses \cite{triplet_super_2018}. This assumption, however, rarely holds in real-world multi-label scenarios where label noise is common.
To tackle noisy supervision, robust methods like CMMQ \cite{CMMQ} and NRCH \cite{NRCH} have been proposed. CMMQ introduces a proxy-based contrastive loss that prioritizes low-loss samples, while NRCH filters out unreliable samples using hard thresholds. However, such filtering risks discarding informative data, and most contrastive learning methods overlook semantic overlap in multi-label scenarios.
These limitations highlight the necessity for more dynamic and noise-tolerant strategies for robust cross-modal hashing.

\section{Method}

\subsection{Problem Definition}

To improve clarity, we first introduce the basic definitions and notations related to cross-modal hashing with noisy labels.
Suppose the multimodal training dataset consists of $N$ examples, denoted as $\mathcal{D} = \left\{ \left\{ \mathbf{x}_i^m \right\}_{m=1}^2, \, \mathbf{y}_i \right\}_{i=1}^N$, where $\mathbf{x}_i^m$ represents the $m$-th modality of the $i$-th sample. Specifically, $m = 1$ corresponds to the image modality and $m = 2$ denotes the text modality. The label vector $\mathbf{y}_i \in \mathbb{R}^C$ is potentially noisy, and $C$ denotes the total number of categories.

Cross-modal hashing learns modality-specific hash functions to embed heterogeneous data into a shared Hamming space. Let $f^m$ denote the hash function for modality $m$, where $m \in \{1, 2\}$. During training, the continuous hash representation $\mathbf{h}_i^m \in \mathbb{R}^L$ is computed as $\mathbf{h}_i^m = \tanh(f^m(\mathbf{x}_i^m))$, where $L$ is the length of the hash codes. The final binary hash codes $\mathbf{b}_i^m \in \{-1, +1\}^L$ are obtained by applying the sign function, i.e., $\mathbf{b}_i^m = \operatorname{sign}(\mathbf{h}_i^m)$.

In addition, a modality-specific linear classifier followed by a sigmoid activation, denoted as $g^m$, is used to produce a category-wise probability distribution. The predicted probability for sample $i$ in modality $m$ is computed as $\mathbf{z}_i^m = g^m(\mathbf{h}_i^m)$, where $\mathbf{z}_i^m \in [0, 1]^C$.

\subsection{Cross-modal Semantic-Consistent Classification}

To fully utilize samples while learning from noisy sample interactions, we exploit cross-modal semantic consistency as a prior to assess label reliability: samples with semantically aligned cross-modal neighbors likely possess accurate labels.
Based on this, we propose the cross-modal semantic-consistent classification (CSCC) module, which adaptively reweights samples by measuring their semantic proximity to neighbors. Firstly, we construct a soft label $\mathbf{p}_i$ for each anchor sample by aggregating the labels of its most similar cross-modal neighbors:
\begin{equation}
\label{eq:neighbor_label}
    \textbf{p}_i =    \sum_{k \in \mathcal{N}_i}  \mathbf{y}_{k} \cdot (\frac{1}{2} \sum_{m=1}^{2}\frac{s_{ik}^m}{\sum_{j \in \mathcal{N}_i} s_{ij}^m} ) \quad ,
\end{equation}
where $s^m_{ik}$ denotes the cosine similarity between the $i$-th sample in modality $m$ and its $k$-th neighbor, $\mathbf{y}_{k}$ represents the label vector of the $k$-th neighbor of sample $i$, and $\mathcal{N}_i$ denotes the set of neighbors of sample $i$.

We then compute a sample-specific confidence weight $w_i$ that reflects how well the current sample's predictions agree with its soft label:

\begin{equation}
\label{eq:weight}
    w_i = \gamma + (1 - \gamma) \cdot \frac{\mathbf{y}_i \cdot \mathbf{p}_i}{\|\mathbf{y}_i\| \cdot \|\mathbf{p}_i\|}  \quad ,
\end{equation}
where $\gamma$ controls the base weight. This formulation measures the semantic consistency between the anchor and its soft label aggregated from cross-modal neighbors. A higher agreement yields a larger weight $w_i$, reflecting stronger supervision confidence and loss contribution.

Finally, we compute the weighted CSCC loss as:

\begin{equation}
\label{eq:classification loss}
\begin{aligned}
{L}_{cscc} = -\frac{1}{2NC}  \sum_{i=1}^N 
    & w_i \sum_{c=1}^C   \sum_{m=1}^2[ z_{ic}^m \log(y_{ic}) \\
    & + (1 - z_{ic}^m) \log(1 - y_{ic}) ] \quad ,
\end{aligned}
\end{equation}
where $\mathbf{z}_{i}^m$ denotes the predicted probability vector over $C$ classes for the $i$-th sample in modality $m$.

This CSCC module enables the model to retain all training samples while attenuating the impact of noisy or misleading labels, thereby providing robust and adaptive supervision in multi-label cross-modal settings.

\subsection{Bidirectional Soft Contrastive Hashing }

Although CSCC leverages clean samples and partially utilizes noisy ones, it overlooks the modality gap and the partial semantic overlaps in multi-label contrastive learning, leading to the loss of critical semantic information.
To address these issues, we propose a novel bidirectional soft contrastive hashing  (BSCH) module.
which comprises two complementary components: an \textbf{Attraction} Component ($L_{att}$) that aligns semantically similar pairs, and a \textbf{Repulsion} Component ($L_{rep}$) that discourages alignment between mismatched samples.
% which adaptively constructs balanced contrastive pairs by leveraging label overlap through a bidirectional strategy shown in Figure \ref{fig:contrastive_strategies}. 
% BSCH consists of two complementary components: an \textbf{Attraction Component} ($L_{att}$) that enhances alignment between semantically similar pairs, and a \textbf{Repulsion Component} ($L_{rep}$) that penalizes spurious alignment between mismatched samples. 

\subsubsection{Construct Contrastive Soft Pairs}

\begin{figure}[t]
    \centering
    \includegraphics[width=0.95\linewidth]{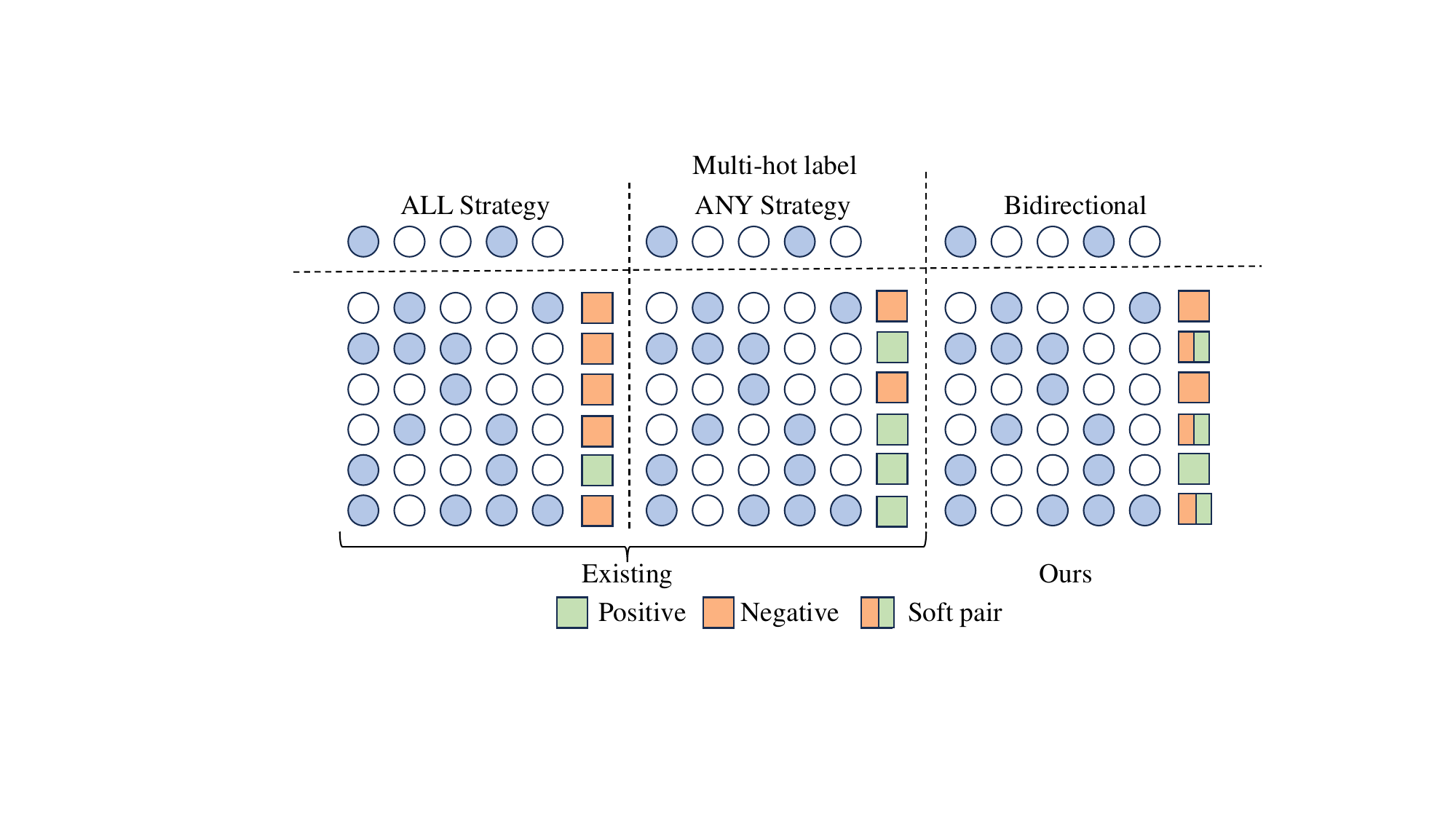} 
    \caption{
    Illustration of three label-based contrastive pairing strategies in multi-label learning. Each row represents the multi-hot label vector of a sample with the first row corresponding to the anchor sample. Blue circle indicates a label is present with a value of one. Green boxes denote positive pairs with the anchor, orange boxes denote negative pairs, and half-green half-orange boxes represent soft pairs.
    Taking the last sample as an example, the ALL strategy constructs it as a negative pair, the ANY strategy regards it as a positive pair, while the Bidirectional strategy treats it as our soft pair due to partial label overlap.
    }
    \label{fig:contrastive_strategies}
\end{figure}
Existing multi-label contrastive methods \cite{contrast1,contrast2} typically adopt either the ALL or ANY strategy (as shown in Figure \ref{fig:contrastive_strategies}) to construct positive and negative pairs: \textbf{ALL} requires identical labels for positives, while \textbf{ANY} considers any shared label as positive. All other pairs are treated as negatives. However, these binary schemes overlook the semantic nuances of partially overlapping labels in multi-label data.

To better reflect these nuanced relationships, we introduce a \textbf{Bidirectional} strategy (Figure \ref{fig:contrastive_strategies}) that categorizes sample pairs as positive (identical labels), negative (no shared labels), or soft (partial overlap), providing a finer-grained view of semantic similarity.
To quantify label similarity using the Jaccard Index. Given two multi-hot label vectors $\mathbf{y}_i, \mathbf{y}_j \in \{0,1\}^C$, the similarity score $R_{ij}$ is defined as:
% \begin{equation}
%      R_{ij} = \frac{ \left| {y}_i \cap {y}_j \right| }{ \left| {y}_i \cup {y}_j \right| }\quad ,
%     \label{eq:jaccard}
% \end{equation}
% where $\cap$ and $\cup$ denote element-wise logical AND and OR operations, respectively, and $|\cdot|$ indicates the number of nonzero entries.
\begin{equation}
    R_{ij} =
    \frac{ \mathbf{{y}}_i^\top \mathbf{{y}}_j }
    { | \mathbf{{y}}_i |_1 + | \mathbf{{y}}_j |_1 - \mathbf{{y}}_i^\top \mathbf{{y}}_j } \quad ,
    \label{eq:jaccard_alt}
\end{equation}
which measures the ratio of shared labels to the total activated labels across both samples.
% The resulting value ranges from 0 to 1, where a higher score indicates stronger semantic alignment.

\subsubsection{Attract}
The attraction loss $L_{att}$ encourages semantic alignment by attracting the similarity of cross-modal positive and soft pairs, and promoting consistency between modalities of the same instance.
\begin{equation}
    L_{att} = \frac{1}{n^2} \sum_{i=1}^n \sum_{ j \neq i}^n \Big[   \exp( \xi - S_{ij})M_{ij} \Big] - \frac{1}{n} \sum_{i=1}^n S_{ii}\quad ,
    \label{contrast_positive}
\end{equation}
where $S_{ij} = \mathbf{h}^1_i \cdot \mathbf{h}^2_j$ measures cross-modal feature similarity between samples $i$ and $j$, and $M_{ij}$ is a binary mask indicating positive pairs with $R_{ij} > 0$, while $S_{ii}$ enforces intra-instance modality alignment as a regularizer. 
The hyperparameter $\xi$ controls the sharpness of positive similarity, guiding the model to perform fine-grained alignment based on label semantics for more coherent cross-modal fusion.

\subsubsection{Repel}
Although soft and negative pairs generally have low label similarity, a subset of them presents high feature similarity, making them difficult to distinguish. To handle these cases, we adopt a margin-based strategy to repel dissimilar pairs with deceptively high similarity: 
\begin{equation}
    N^{d}_{ij} = S^{d}_{ij} - \xi \cdot max(0,(S_{ii} - m)-S^{d}_{ij}) ,
\end{equation}
where $S^{d}_{ij}$ denotes the raw cross-modal similarity in direction $d$, with $S^{12}_{ij} = S_{ij}$ for image-to-text and $S^{21}_{ij} = S_{ji}$ for text-to-image retrieval. The hyperparameter $m$ serves as a margin threshold to distinguish hard negatives, while $\xi$ is a decay term used to down-weight weakly aligned pairs.

% 顺序
Then the repulsion loss $L_{rep}$ is defined as:
\begin{equation}
    L_{rep} = \frac{1}{2n^2} \sum_{i=1}^n \sum_{ j \neq i}^n \left[\sum^2_{d=1} \exp\left(  N^{d}_{ij}(1 - R_{ij})\right) \right] \quad .
    \label{contrast_negative}
\end{equation}
This component repels high feature similarity for semantically dissimilar pairs, with adaptive weighting: high label similarity $R_{ij}$ reduces the penalty, while high feature similarity $N^d_{ij}$ under low label agreement increases it.

The final bidirectional soft contrastive hashing loss integrates both attraction and repulsion components, along with a regularization term:
\begin{equation}
    L_{bsch} = L_{att} + L_{rep}  + \frac{\beta}{n\cdot L} \sum^2_{m=1}\sum^n_{i=1}\sum^L_{l=1}( \left|h^m_{il}\right| - 1 )  ,
    \label{eq:bidirectional_contrast}
\end{equation}
the last term acts as a regularizer, encouraging each element of the hash codes $h^m_{il}$ to approach binary values, with $\beta$ as a hyperparameter governing the regularization strength.

This bidirectional attraction-repulsion design allows a sample pair to be treated as soft pairs depending on label overlap. It pulls semantically aligned instances closer, while pushing mismatched ones apart. This mechanism enhances the robustness and discriminability of hash codes in noisy multi-label cross-modal scenarios.
% COCO NUS MAP Table
\begin{table*}[ht]
\renewcommand{\arraystretch}{0.86} % 调整行高
\setlength{\tabcolsep}{1.5mm}
\begin{tabular}{c c c | cccc | cccc | cccc}
\toprule

\multirow{2}{*}{\centering \textbf{Dataset}} & \multirow{2}{*}{\centering \textbf{Method}} & \multirow{2}{*}{\centering \textbf{Year}} & \multicolumn{4}{c|}{\textbf{20\%}} & \multicolumn{4}{c|}{\textbf{50\%}} & \multicolumn{4}{c}{\textbf{80\%}} \\

& & & 16bit & 32bit & 64bit & 128bit & 16bit & 32bit & 64bit & 128bit & 16bit & 32bit & 64bit & 128bit \\
\midrule
\multirow{13}{*}{WIDE}
& DJSRH &2019 & 42.7 & 46.5 & 49.7 & 52.4 & 42.7 & 46.5 & 49.7 & 52.4 & 42.7 & 46.5 & 49.7 & 52.4 \\
& DGCPN &2021 & 59.2 & 61.5 & 63.3 & 64.4 & 59.2 & 61.5 & 63.3 & 64.4 & 59.2 & 61.5 & 63.3 & 64.4 \\
& PIP   &2021 & 56.7 & 58.1 & 59.7 & 60.2 & 56.7 & 58.1 & 59.7 & 60.2 & 56.7 & 58.1 & 59.7 & 60.2 \\
& CIRH  &2022 & 56.3 & 58.6 & 59.3 & 60.4 & 56.3 & 58.6 & 59.3 & 60.4 & 56.3 & 58.6 & 59.3 & 60.4 \\
& UCCH  &2023 & 60.3 & 62.7 & 63.7 & 64.4 & 60.3 & 62.7 & 63.7 & 64.4 & {60.3} &  \underline{62.7} & 63.7 & 64.4 \\

\cmidrule{2-15}

& CPAH  &2020 & 63.3 & 66.5 & 67.7 & 67.9 & 57.3 & 60.5 & 61.3 & 60.5 & 49.2 & 52.9 & 54.7 & 53.8 \\
& CMMQ  &2022 & 62.2 & 63.4 & 64.2 & 64.7 & 59.7 & 60.7 & 61.5 & 61.9 & 50.9 & 50.6 & 51.7 & 52.4 \\
& MIAN  &2023 & 60.3 & 61.9 & 63.2 & 63.5 & 54.2 & 56.5 & 57.8 & 58.0 & 46.8 & 48.2 & 49.6 & 48.6 \\
& LtCMH &2023 & 59.4 & 60.5 & 62.7 & 63.5 & 49.3 & 49.8 & 50.7 & 51.7 & 44.6 & 43.8 & 45.6 & 45.8 \\
& NRCH  &2024 & \underline{65.3} &  \underline{67.6} &  \underline{69.1} &  \underline{68.8} & 63.1 & 65.1 &  \underline{67.0} &  \underline{68.1} &  \underline{61.0} & 62.3 &  \underline{63.9} &  \underline{64.7} \\
& DHRL  &2024 & 57.6 & 61.6 & 61.6 & 64.2 & 52.8 & 54.3 & 56.7 & 61.6 & 32.0 & 40.4 & 50.8 & 56.0 \\
& RSHNL &2025 & 63.3 & 65.3 & 66.5 & 67.5 &  \underline{63.3} &  \underline{65.3} & 66.0 & 67.2 & 57.8 & 59.9 & 62.6 & 64.1 \\
& \textbf{Our SCBCH} & & \textbf{66.7} & \textbf{68.2} & \textbf{69.6} & \textbf{69.7} & \textbf{63.9} & \textbf{66.4} & \textbf{67.5} & \textbf{68.1} & \textbf{62.0} & \textbf{63.9} & \textbf{65.0} & \textbf{66.0} \\
\midrule
\multirow{13}{*}{COCO}
& DJSRH &2019 & 47.8 & 50.6 & 54.4 & 57.1 & 47.8 & 50.6 & 54.4 & 57.1 & 47.8 & 50.6 & 54.4 & 57.1 \\
& DGCPN &2021 & 60.5 & 63.2 & 63.1 & 64.4 & 60.5 & 63.2 & 63.1 & 64.4 & 60.5 & 63.2 & 63.1 & 64.4 \\
& PIP   &2021 & 54.2 & 56.2 & 57.7 & 59.4 & 54.2 & 56.2 & 57.7 & 59.4 & 54.2 & 56.2 & 57.7 & 59.4 \\
& CIRH  &2022 & 61.1 & 63.1 & 64.0 & 64.1 & 61.1 & 63.1 & 64.0 & 64.1 & 61.1 & 63.1 & 64.0 & 64.1 \\
& UCCH  &2023 & 55.3 & 61.4 & 61.2 & 62.5 & 55.3 & 61.4 & 61.2 & 62.5 & 55.3 & 61.4 & 61.2 & 62.5 \\

\cmidrule{2-15}

& CPAH  &2020 &  \underline{64.7} &  \underline{68.2} &  \underline{68.8} &  \underline{69.2} & 61.3 & 66.2 &  \underline{68.5} &  \underline{69.0} & 57.4 & 64.7 & 66.4 & 64.8 \\
& CMMQ  &2022 & 56.8 & 58.6 & 60.3 & 61.0 & 54.3 & 56.2 & 58.0 & 57.7 & 49.7 & 50.4 & 52.8 & 53.1 \\
& MIAN  &2023 & 58.7 & 61.4 & 62.3 & 63.1 & 56.9 & 57.6 & 59.0 & 59.8 & 52.0 & 56.0 & 57.0 & 57.5 \\
& LtCMH &2023 & 55.8 & 59.7 & 62.6 & 65.3 & 53.1 & 56.6 & 57.9 & 56.3 & 51.7 & 52.9 & 55.2 & 58.0 \\
& NRCH  &2024 & 63.5 & 68.0 & 67.5 & 67.8 &  \underline{65.4} &  \underline{68.2} & 68.4 & 68.6 &  \underline{63.9} &  \underline{67.1} &  \underline{68.3} &  \underline{69.0} \\
& DHRL  &2024 & 33.8 & 33.4 & 45.4 & 61.4 & 33.8 & 33.4 & 57.3 & 60.7 & 33.8 & 33.4 & 49.2 & 58.1 \\
& RSHNL &2025 & 58.2 & 61.1 & 63.4 & 64.0 & 59.4 & 61.3 & 64.4 & 64.4 & 59.3 & 61.9 & 63.9 & 64.8 \\

& \textbf{Our SCBCH} & & \textbf{67.2} & \textbf{68.8} & \textbf{69.7} & \textbf{70.6} & \textbf{66.8} & \textbf{68.2} & \textbf{70.0} & \textbf{70.4} & \textbf{65.2} & \textbf{67.7} & \textbf{69.7} & \textbf{69.5} \\
\bottomrule
\end{tabular}%
\caption{The performance comparison in terms of average MAP scores of I2T and T2I tasks on the NUS-WIDE (NUS) and MS-COCO (COCO) datasets is presented, with highest and the second-highest scores are in \textbf{bold} and \underline{underlined}, respectively.}

\label{tab:nus_coco_map_result}
\end{table*}

\subsection{Optimization}

By combining the above loss terms, the final objective function of our model is defined as:
\begin{equation}
    \label{eq:final_loss}
    \begin{aligned}
    L &= 
        \begin{cases}
            L_{c} + \alpha L_{bsch}       & \text{if } t \leq E_w \\
            L_{cscc} + \alpha L_{bsch},       & \text{if } E_w<t<E_m
        \end{cases}  \quad ,
\end{aligned}
\end{equation}
where $\alpha$ is a hyperparameter, $t$ is the current epoch, $E_w$ is the warm-up epoch, and $E_m$ is the total training epochs. For $t \leq E_w$, we disable the semantic-consistency weighting strategy and use unweighted classification loss $L_{c}$ by setting $w_i = 1$ in $L_{cscc}$. Once $t > E_w$, we switch to $L_{cscc}$ with confidence-based weights for adaptive supervision.

\section{Experiments}

% 图数据集名称
% FILCKR IAPR MAP Table
\begin{table*}[ht]
\renewcommand{\arraystretch}{0.86} % 调整行高
\setlength{\tabcolsep}{1.5mm}
\begin{tabular}{c c c | cccc | cccc | cccc}
\toprule

\multirow{2}{*}{\centering \textbf{Dataset}} & \multirow{2}{*}{\centering \textbf{Method}} & \multirow{2}{*}{\centering \textbf{Year}} & \multicolumn{4}{c|}{\textbf{20\%}} & \multicolumn{4}{c|}{\textbf{50\%}} & \multicolumn{4}{c}{\textbf{80\%}} \\

& & & 16bit & 32bit & 64bit & 128bit & 16bit & 32bit & 64bit & 128bit & 16bit & 32bit & 64bit & 128bit \\
\midrule
\multirow{13}{*}{Flickr}
& DJSRH &2019 & 62.5 & 63.6 & 65.2 & 66.7 & 62.5 & 63.6 & 65.2 & 66.7 & 62.5 & 63.6 & 65.2 & 66.7 \\
& DGCPN &2021 & 68.9 & 69.5 & 70.7 & 71.3 & 68.9 & 69.5 & 70.7 & 71.3 & 68.9 & 69.5 & 70.7 & 71.3 \\
& PIP   &2021 & 68.0 & 68.0 & 69.3 & 70.0 & 68.0 & 68.0 & 69.3 & 70.0 & 68.0 & 68.0 & 69.3 & 70.0 \\
& CIRH  &2022 & 68.0 & 68.4 & 69.3 & 69.9 & 68.0 & 68.4 & 69.3 & 69.9 & 68.0 & 68.4 & 69.3 & 69.9 \\
& UCCH  &2023 & 70.1 & 71.4 & 72.0 & 72.2 & 70.1 & 71.4 & 72.0 & 72.2 & 70.1 & 71.4 & 72.0 & 72.2 \\

\cmidrule{2-15}

& CPAH  &2020 &  \underline{76.9} &  \underline{78.6} &  \underline{79.2} &  \underline{79.2} &  \underline{75.0} &  \underline{76.9} &  \underline{77.4} & 76.5 &  \underline{74.1} &  \underline{75.0} & 75.7 & 74.6 \\
& CMMQ  &2022 & 68.0 & 71.2 & 73.1 & 73.8 & 66.1 & 70.0 & 71.4 & 72.1 & 63.3 & 67.5 & 68.2 & 69.2 \\
& MIAN  &2023 & 76.1 & 77.3 & 78.1 & 78.3 & 74.4 & 75.2 & 76.2 &  \underline{76.6} & 71.5 & 73.4 & 74.4 & 74.9 \\
& LtCMH &2023 & 61.5 & 75.0 & 75.7 & 76.3 & 68.9 & 72.4 & 73.9 & 74.2 & 68.4 & 71.0 & 71.6 & 72.0 \\
& NRCH  &2024 & 75.8 & 76.0 & 76.5 & 76.4 & 74.7 & 75.3 & 76.2 & 76.0 & 73.7 & 74.3 &  \underline{75.8} &  \underline{76.0} \\
& DHRL  &2024 & 69.0 & 71.2 & 73.2 & 71.5 & 67.3 & 70.7 & 71.3 & 70.4 & 66.1 & 67.7 & 70.5 & 69.0 \\
& RSHNL &2025 & 73.0 & 74.2 & 75.5 & 75.9 & 71.9 & 73.1 & 74.2 & 74.8 & 71.2 & 72.4 & 74.0 & 73.9 \\
& \textbf{Our SCBCH} & & \textbf{78.0} & \textbf{79.2} & \textbf{79.7} & \textbf{79.8} & \textbf{76.2} & \textbf{77.9} & \textbf{78.6} & \textbf{78.8} & \textbf{74.7} & \textbf{76.1} & \textbf{77.4} & \textbf{77.9} \\
\midrule
\multirow{13}{*}{IAPR}
& DJSRH &2019 & 36.7 & 38.8 & 41.4 & 43.3 & 36.7 & 38.8 & 41.4 & 43.3 & 36.7 & 38.8 & 41.4 & 43.3 \\
& DGCPN &2021 & 46.1 & 47.6 & 48.8 & 49.6 & 46.1 & 47.6 & 48.8 & 49.6 & 46.1 & 47.6 & 48.8 & 49.6 \\
& PIP   &2021 & 45.1 & 46.6 & 47.7 & 48.4 & 45.1 & 46.6 & 47.7 & 48.4 & 45.1 & 46.6 & 47.7 & 48.4 \\
& CIRH  &2022 & 44.4 & 46.3 & 47.8 & 49.0 & 44.4 & 46.3 & 47.8 & 49.0 & 44.4 & 46.3 & 47.8 & 49.0 \\
& UCCH  &2023 & 44.7 & 46.1 & 47.1 & 47.6 & 44.7 & 46.1 & 47.1 & 47.6 & 44.7 & 46.1 & 47.1 & 47.6 \\

\cmidrule{2-15}

& CPAH  &2020 &  \underline{51.0} &  \underline{53.6} & 54.9 & 55.6 &  \underline{49.9} & 52.2 & 53.0 & 54.0 & 48.5 & 50.7 & 51.4 & 52.2 \\
& CMMQ  &2022 & 38.3 & 39.0 & 41.7 & 43.3 & 38.1 & 38.6 & 41.3 & 42.9 & 37.5 & 37.6 & 40.3 & 41.7 \\
& MIAN  &2023 & 46.2 & 48.7 & 50.2 & 51.5 & 44.8 & 46.8 & 48.3 & 49.5 & 43.0 & 45.4 & 46.1 & 47.4 \\
& LtCMH &2023 & 44.4 & 45.8 & 47.6 & 48.6 & 44.2 & 45.6 & 47.2 & 47.8 & 44.0 & 45.1 & 46.6 & 47.3 \\
& NRCH  &2024 & 50.7 & 53.1 &  \underline{55.0} &  \underline{55.7} &  \underline{49.9} &  \underline{52.8} &  \underline{54.4} &  \underline{55.5} &  \underline{49.2} &  \underline{52.0} &  \underline{53.8} &  \underline{55.0} \\
& DHRL  &2024 & 30.0 & 30.1 & 30.5 & 43.8 & 30.0 & 30.1 & 30.5 & 43.1 & 30.0 & 30.1 & 30.5 & 42.3 \\
& RSHNL &2025 & 44.4 & 45.5 & 45.7 & 46.0 & 43.8 & 45.8 & 45.9 & 45.8 & 44.2 & 44.5 & 45.3 & 45.5 \\

& \textbf{Our SCBCH} & & \textbf{53.0} & \textbf{55.6} & \textbf{58.1} & \textbf{59.2} & \textbf{52.0} & \textbf{54.5} & \textbf{57.3} & \textbf{58.3} & \textbf{50.7} & \textbf{53.8} & \textbf{56.1} & \textbf{57.2} \\
\bottomrule
\end{tabular}%

\caption{The performance comparison in terms of average MAP scores of I2T and T2I tasks on the MIRFlickr-25K (Flickr) and IAPR TC-12 (IAPR) datasets is presented, with highest and the second-highest scores are in \textbf{bold} and \underline{underlined}, respectively.}

\label{tab:flickr_iapr_map_result}

\end{table*}

% PR CURVE
\begin{figure*}[!ht]
    \centering
    \begin{subfigure}[t]{0.22\textwidth}
        \includegraphics[width=\linewidth]{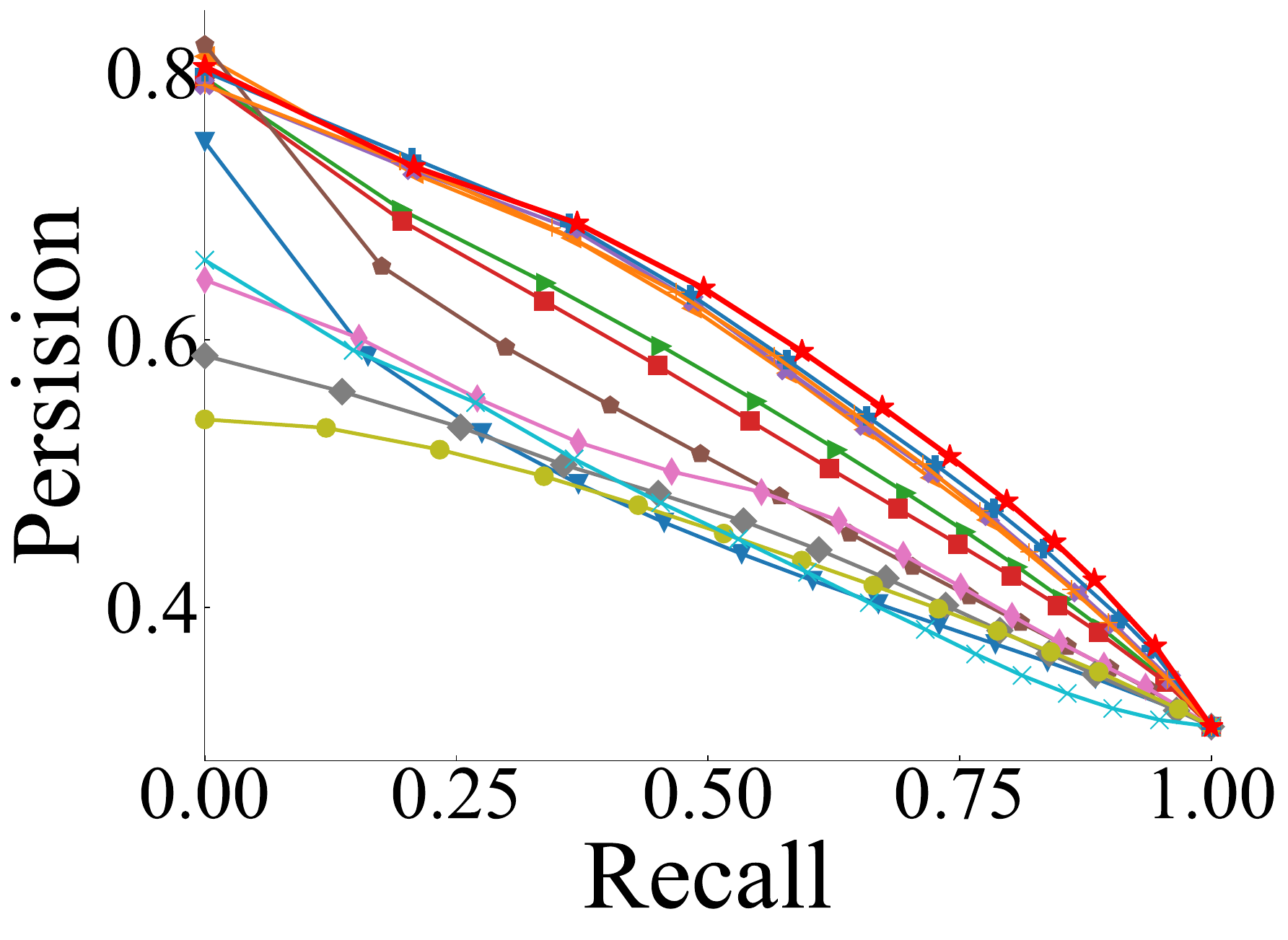}
        \caption{NUS (I2T)}
    \end{subfigure}
    \hfill
    \begin{subfigure}[t]{0.22\textwidth}
        \includegraphics[width=\linewidth]{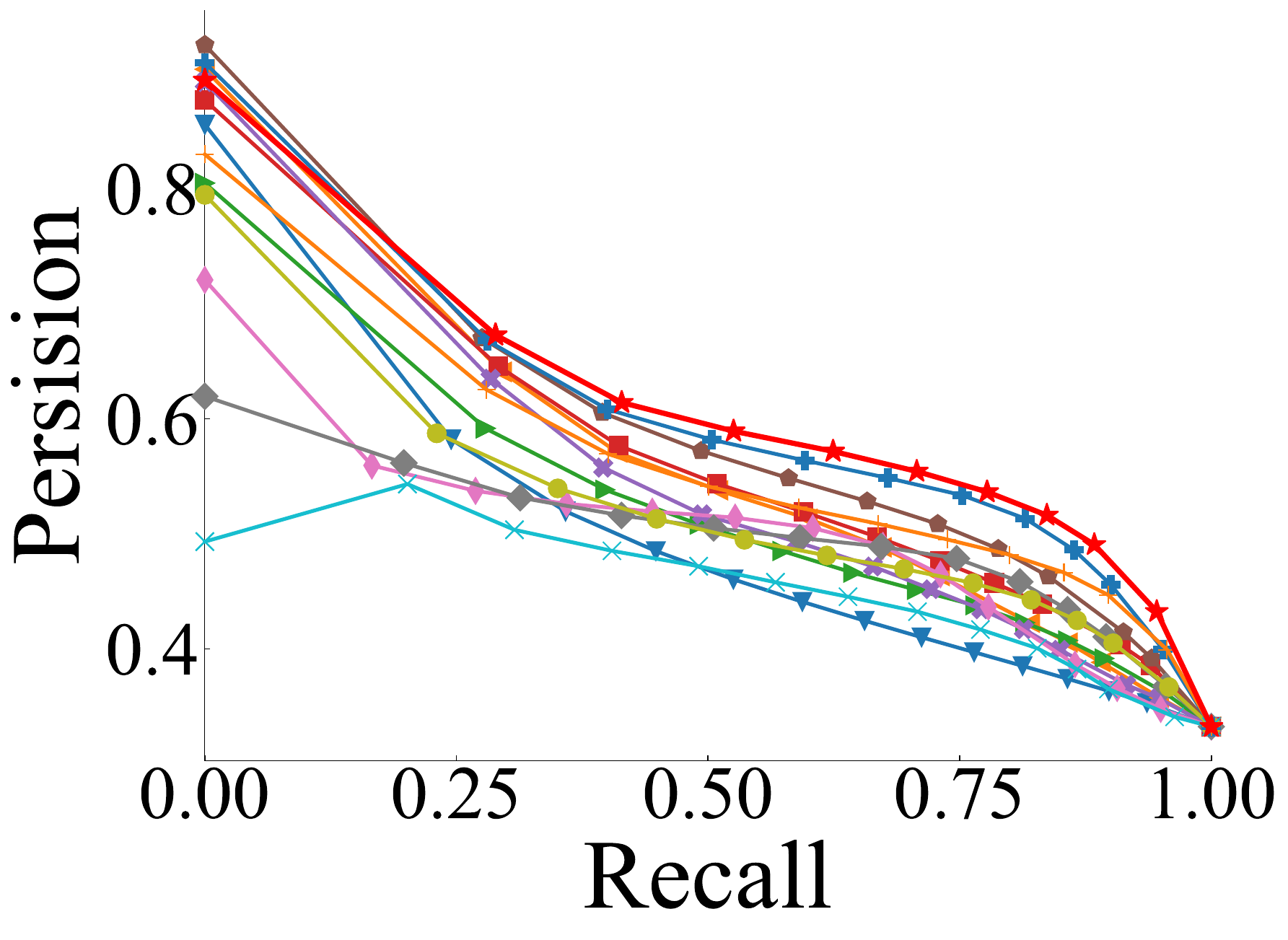}
        \caption{COCO (I2T)}
    \end{subfigure}
    \hfill
    \begin{subfigure}[t]{0.22\textwidth}
        \includegraphics[width=\linewidth]{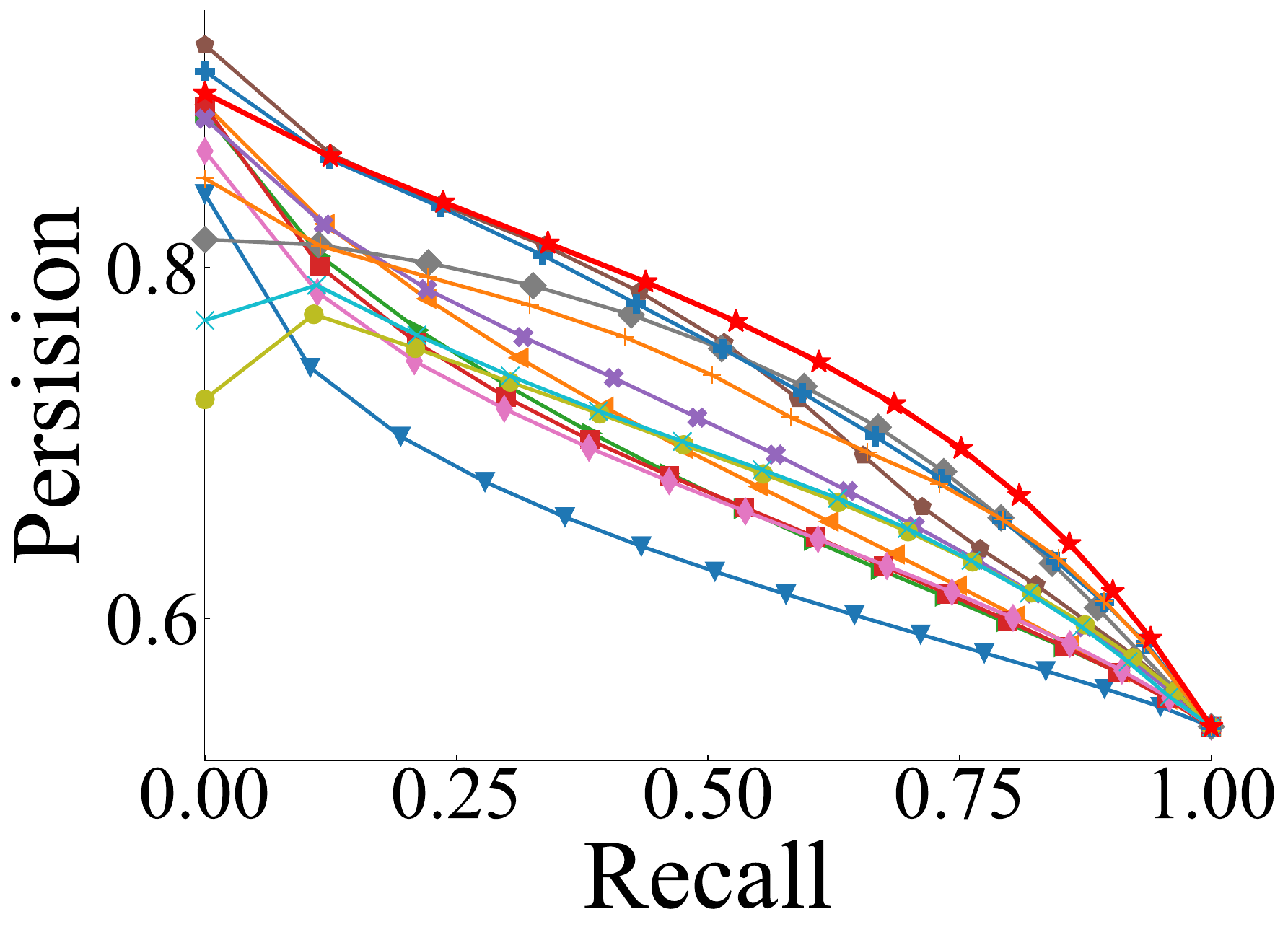}
        \caption{Flickr (I2T)}
    \end{subfigure}
    \hfill
    \begin{subfigure}[t]{0.27\textwidth}
        \includegraphics[width=\linewidth]{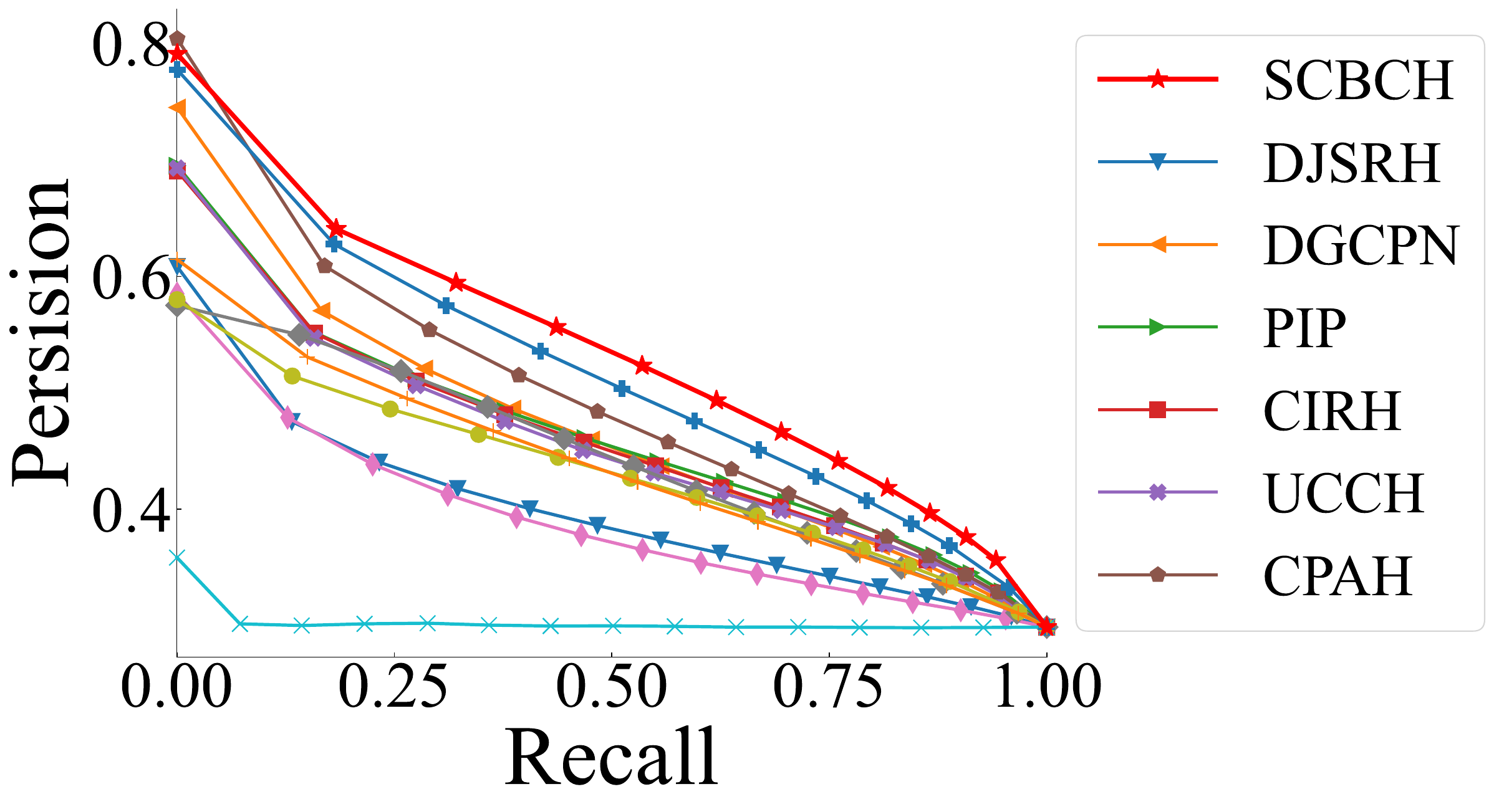}
        \caption{IAPR (I2T)}
    \end{subfigure}

    \begin{subfigure}[t]{0.22\textwidth}
        \includegraphics[width=\linewidth]{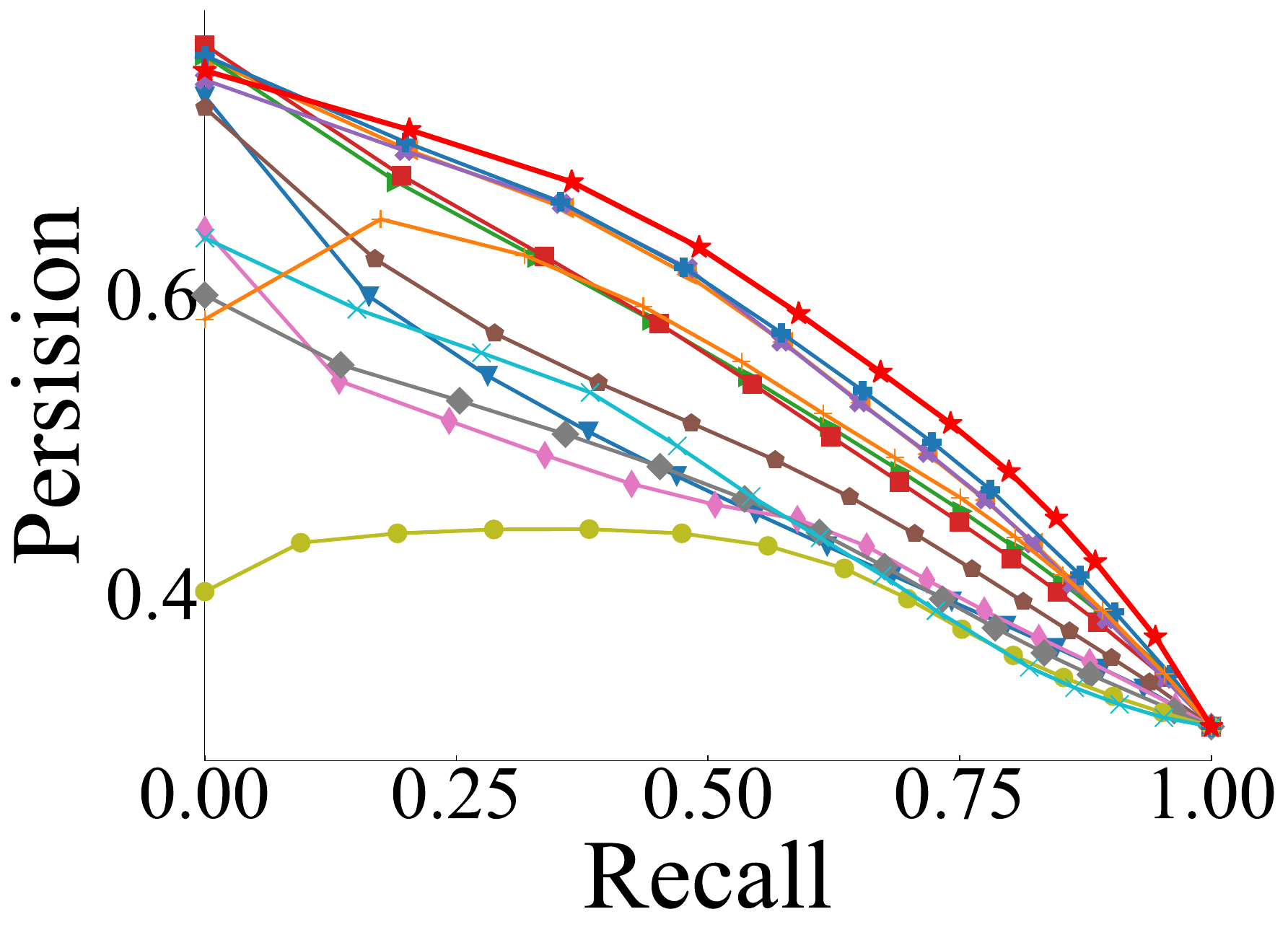}
        \caption{NUS (T2I)}
    \end{subfigure}
    \hfill
    \begin{subfigure}[t]{0.22\textwidth}
        \includegraphics[width=\linewidth]{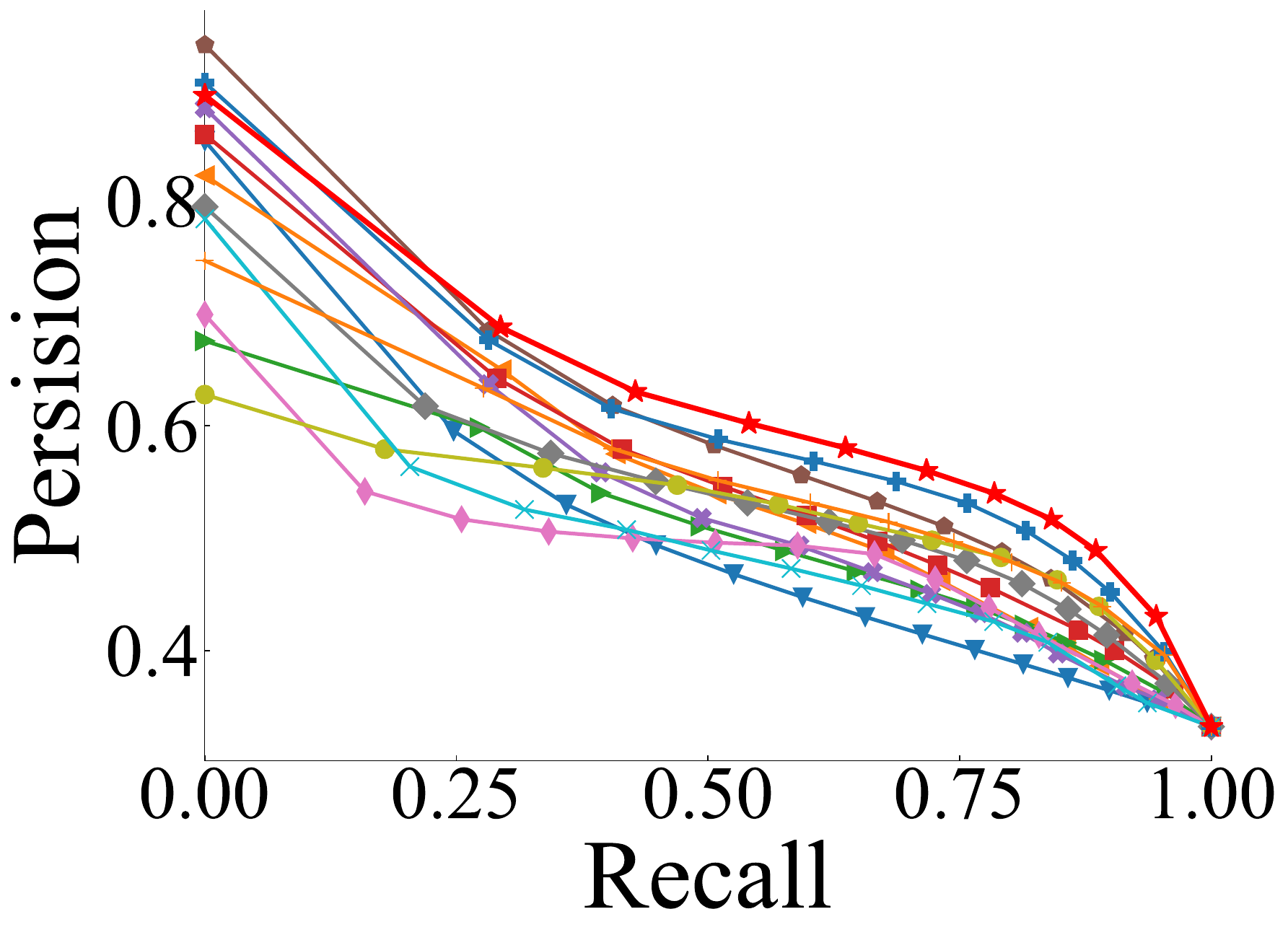}
        \caption{COCO (T2I)}
    \end{subfigure}
    \hfill
    \begin{subfigure}[t]{0.22\textwidth}
        \includegraphics[width=\linewidth]{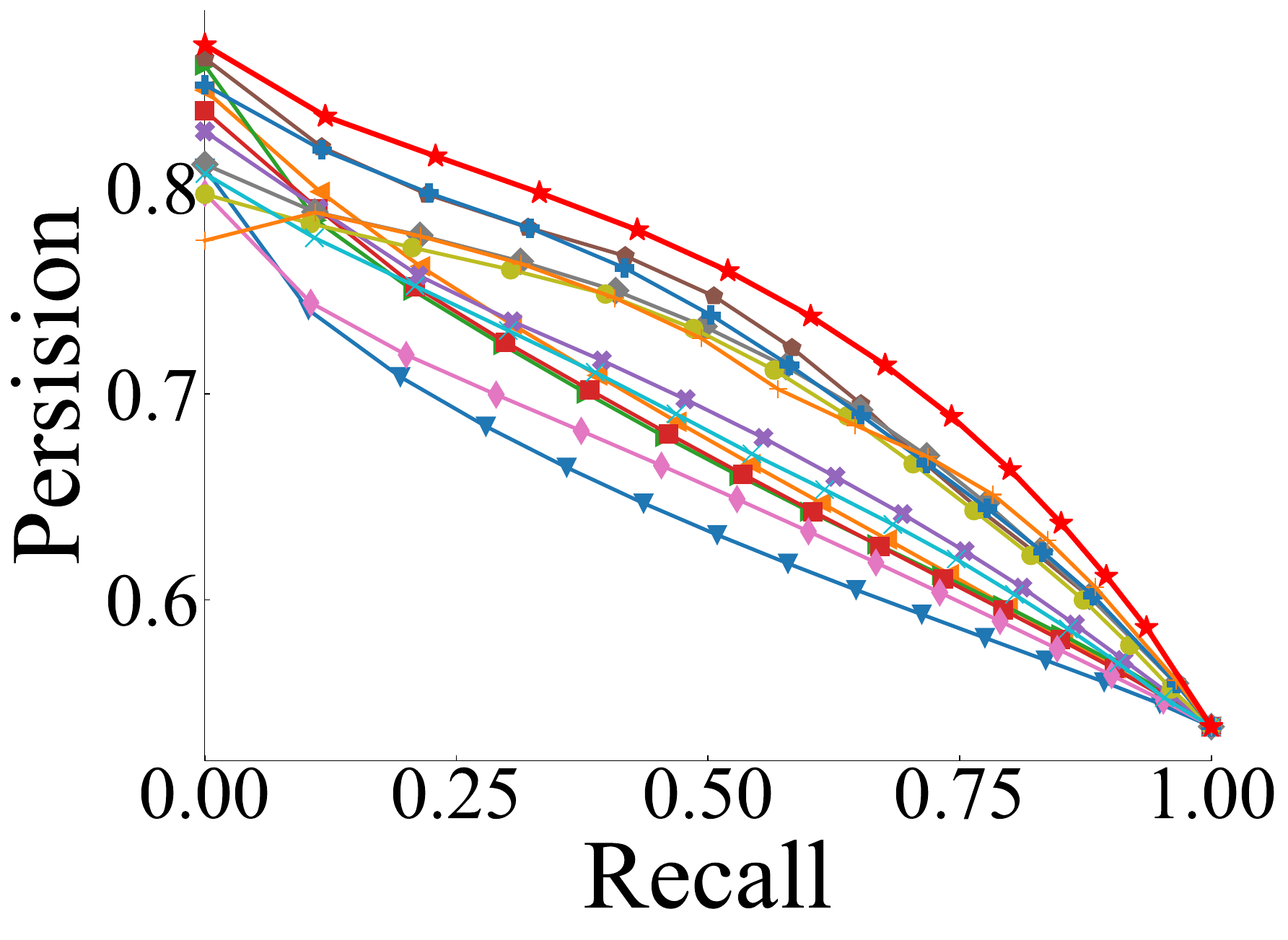}
        \caption{Flickr (T2I)}
    \end{subfigure}
    \hfill
    \begin{subfigure}[t]{0.27\textwidth}
        \includegraphics[width=\linewidth]{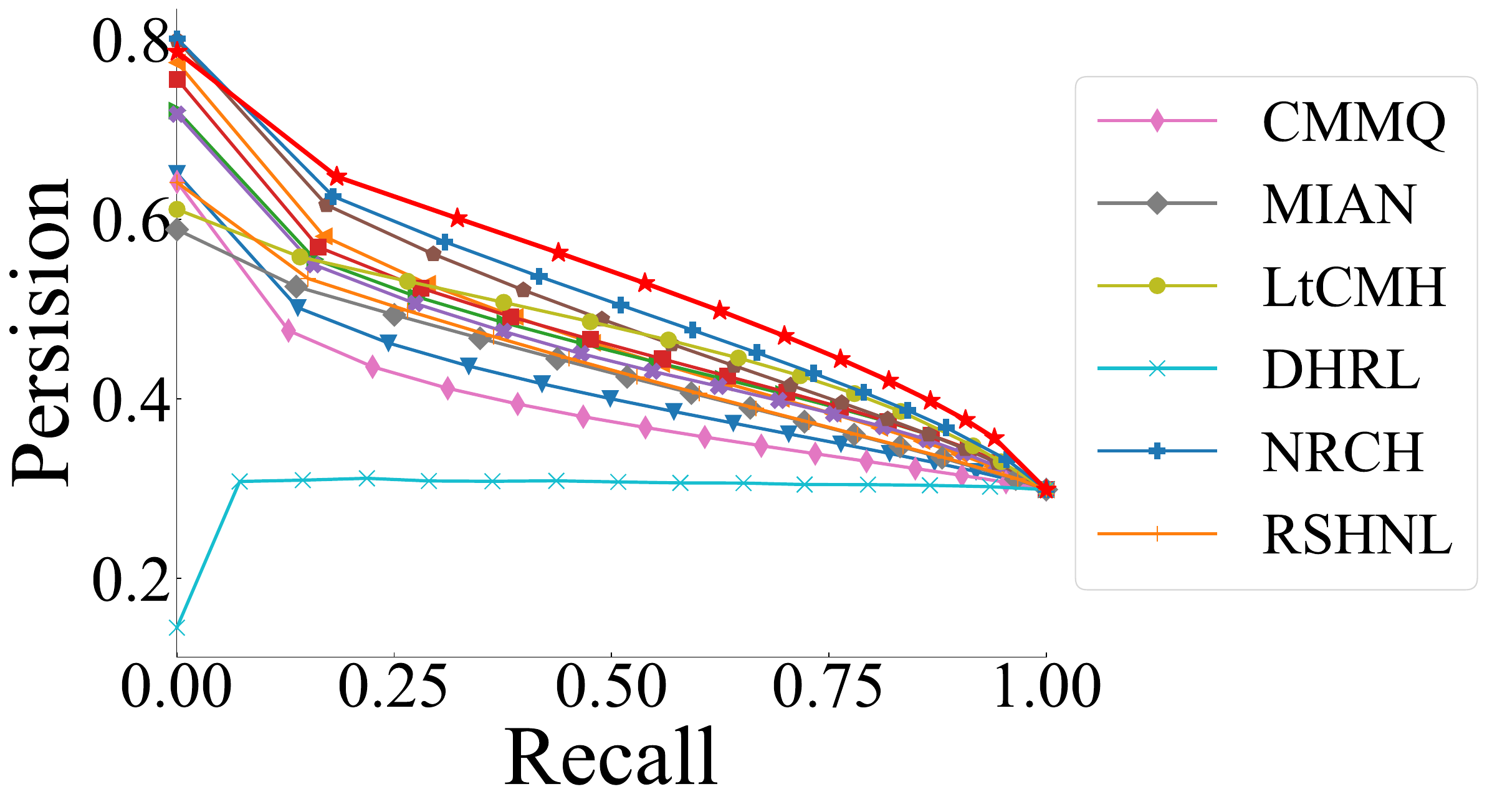}
        \caption{IAPR (T2I)}
    \end{subfigure}

    \caption{Precision-recall curves on four datasets under 64-bit hash codes and a 0.8 noise rate.}
    \label{fig:PR}
\end{figure*}

\subsection{Experimental Setup}
To evaluate the performance of our method, we conducted extensive experiments across four widely-used benchmark datasets, \textbf{NUS-WIDE} \cite{dataset_nus}, \textbf{MS-COCO} \cite{dataset_coco}, \textbf{MIRFlickr-25K} \cite{dataset_flickr}, and \textbf{IAPR TC-12} \cite{dataset_iapr}.
And we conduct two cross-modal retrieval tasks: image-to-text (I2T) and text-to-image (T2I), where retrieval is based on Hamming distance. Similar to \cite{hu2022UCCH}, we adopt the widely used Mean Average Precision (MAP) as the evaluation metric. MAP is computed as the mean of Average Precision (AP) scores over all queries, effectively reflecting both the precision and ranking quality of the retrieved results. MAP scores are reported under different hash code lengths (16, 32, 64, and 128 bits). To assess robustness, we inject mixed symmetric label noise \cite{symmetric_label_noise} into the training data at noise rates of 20\%, 50\%, and 80\%.

\subsection{Implementation Details}
In SCBCH, we use the pre-trained VGG19~\cite{vgg19} on ImageNet for image features and Doc2Vec~\cite{doc2vec} for text features. Three FC layers are added to the image branch and two to the text branch, each with ReLU activations except the last. Hidden layers have 8,192 units, and outputs are projected to dimension $L$ for hash codes.
Training employs Adam~\cite{Adam} with a learning rate of $1\times10^{-4}$, batch size 128, and 50 epochs. Additional specific parameters are set to $\gamma=0.5$, $\alpha=0.7$, and $\beta=0.3$. To ensure a fair comparison, all backbone networks are kept frozen during training. Our SCBCH is implemented using the PyTorch framework and trained on a single GeForce RTX3090 24GB GPU.

\subsection{Comparison with State-of-the-Arts}

% Contrastive Similarity Analysis figure
\begin{figure*}[ht]
    \centering
    \begin{subfigure}[t]{0.29\textwidth}
        \centering
        \includegraphics[width=\linewidth]{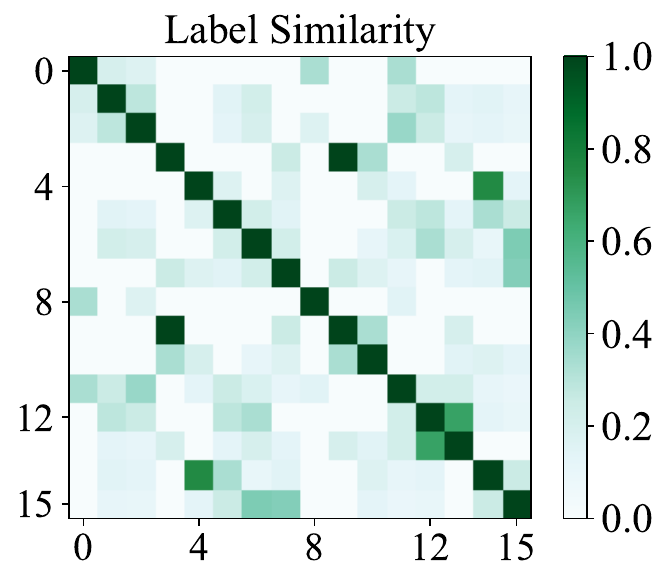}
        \caption{Label similarity}
    \end{subfigure}
    \hfill
    \begin{subfigure}[t]{0.3\textwidth}
        \centering
        \includegraphics[width=\linewidth]{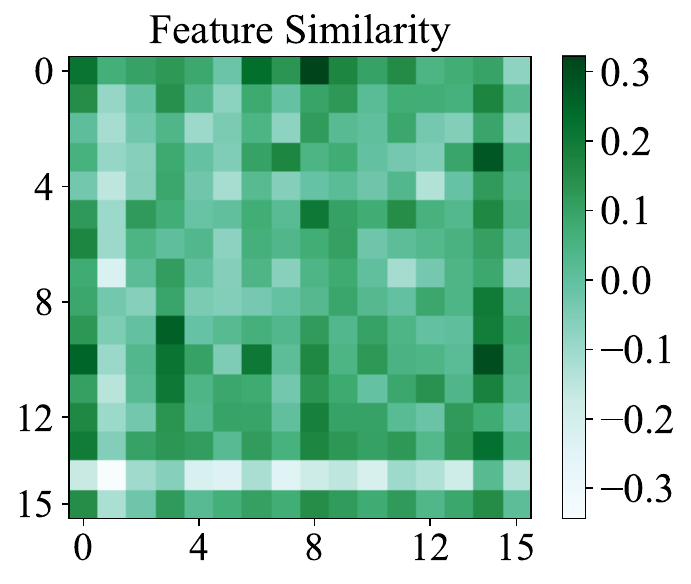}
        \caption{Initial feature similarity (epoch 1)}
    \end{subfigure}
    \hfill
    \begin{subfigure}[t]{0.3\textwidth}
        \centering
        \includegraphics[width=\linewidth]{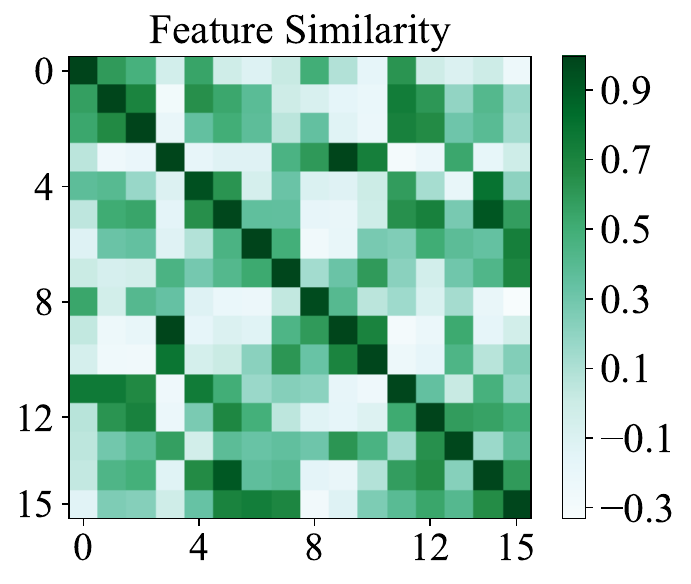}
        \caption{Final feature similarity (epoch 50)}
    \end{subfigure}

    \caption{Heatmaps of label similarities and cross-modal feature similarities among 16 randomly selected samples on the MIRFlickr-25K dataset under a 50\% noise rate. Subfigure (a) shows the label similarity matrix. Subfigures (b) and (c) illustrate the evolution of feature similarities from epoch 1 to epoch 50.}
    \label{fig:similarity}
\end{figure*}

To validate the effectiveness of our proposed method, we conduct comprehensive comparisons against 12 state-of-the-art methods on four widely adopted cross-modal benchmark datasets. These baselines include five unsupervised approaches: DJSRH \cite{DJSRH}, DGCPN \cite{DGCPN}, PIP \cite{PIP}, CIRH \cite{unsup_CIRH_zhu2022work}, and UCCH \cite{hu2022UCCH}, as well as seven supervised methods: CPAH \cite{CPAH_2020}, CMMQ \cite{CMMQ}, MIAN \cite{sup_MIAN_TKDE2023}, LtCMH \cite{sup_LtCMH_AAAI2023}, NRCH \cite{NRCH}, DHRL \cite{sup_DHRL_TBD2024}, and RSHNL \cite{sup_RSHNL_AAAI2025}.
For each method, we track performance across training epochs and report the best MAP scores. Tables \ref{tab:nus_coco_map_result} and \ref{tab:flickr_iapr_map_result} report the average MAP scores for both I2T and T2I retrieval tasks. Additionally, Figure \ref{fig:PR} plots the precision-recall curves under a 64-bit code length and 80\% noise setting.
Based on the results in Tables \ref{tab:nus_coco_map_result} and \ref{tab:flickr_iapr_map_result}, we draw the following key observations:

\begin{itemize}
    % \item Longer hash codes generally boost performance via enhanced semantic encoding, yet supervised methods degrade with noise while unsupervised counterparts remain noise-resilient yet improvement-constrained by unlabeled training.
    \item Longer hash codes generally boost performance via enhanced semantic encoding, but supervised methods suffer under noise, whereas unsupervised ones resist noise yet show limited gains without labels.

    \item As shown in Tables \ref{tab:nus_coco_map_result} and \ref{tab:flickr_iapr_map_result}, the proposed SCBCH consistently outperforms all competing methods. For example, with a hash code length of 64 and a noise rate of 0.8, SCBCH achieves performance gains of 1.1\%, 1.4\%, 1.6\%, and 2.3\% over the second-best method on the NUS-WIDE, MS-COCO, MIRFlickr-25K, and IAPR TC-12 datasets, respectively.

    \item As shown in Figure \ref{fig:PR}, SCBCH consistently achieves the highest area under the precision-recall curves for both I2T and T2I tasks, demonstrating its stable and superior performance compared to state-of-the-art methods.
    
\end{itemize}

\subsection{Ablation Study}
To thoroughly investigate the contribution of each component, we compare SCBCH with its four variants: (1) removing the CSCC loss $L_{cscc}$, (2) removing the BSCH loss $L_{bsch}$, (3) removing the semantic-consistency weighting strategy, and (4) removing the attraction component loss $L_{att}$.

To ensure fair comparison, all variants are trained under the same configuration as SCBCH and evaluated on the test datasets. As shown in Table \ref{tab:ablation}, the complete SCBCH achieves the best performance, while the variants yield inferior results. This confirms that each component plays a critical role in the overall effectiveness of SCBCH.

% Ablation table
\begin{table}[t]
    \renewcommand{\arraystretch}{0.86} % 调整行高

    \centering
    \begin{tabular}{c | ccc | ccc}
    \toprule
    Dataset & \multicolumn{3}{c|}{\textbf{COCO}} & \multicolumn{3}{c}{\textbf{IAPR}} \\
    \midrule
    Noise & 0.2 & 0.5 & 0.8 & 0.2 & 0.5 & 0.8\\
    \midrule
    
    SCBCH-1  &  68.0 & 68.2 &  68.6 & 57.3 &  56.1 & 55.3 \\
    SCBCH-2  &  36.8 & 37.0 &  36.0 &  30.4&  30.4 & 30.4 \\
    SCBCH-3  &  69.4 & 69.5 &  69.3 & 57.9 &  56.9 & 55.8 \\
    SCBCH-4  &  65.9 & 64.6 &  64.2 & 47.2 &  47.1 & 46.9 \\
    SCBCH  &  \textbf{69.7} & \textbf{70.0} &  \textbf{69.7} & \textbf{58.1} &  \textbf{57.3} & \textbf{56.1} \\
    \bottomrule
    \end{tabular}
    \caption{Ablation results on the MS-COCO and IAPR TC-12 dataset with 64-bit codes under all noise rates are presented.}
    
    \label{tab:ablation}
\end{table}

% \subsection{Parameter Sensitivity Analysis}

\subsection{Model Analysis}

\subsubsection{Contrastive Similarity Analysis}

Figure~\ref{fig:similarity} shows label and cross-modal feature similarity matrices for 16 MIRFlickr-25K samples under 50\% noise with 64-bit codes. Subfigure (a) displays the label similarity heatmap, capturing partial semantic overlaps among samples. Subfigures (b) and (c) show the feature similarity matrices at epoch 1 and epoch 50, respectively. Initially disordered, they become well aligned with labels after training, demonstrating the effectiveness of our bidirectional contrastive strategy in enhancing noise-robust semantic consistency.

\begin{figure}[t]
    \centering
    \begin{subfigure}[t]{0.48\linewidth}
        \centering
        \includegraphics[width=\linewidth]{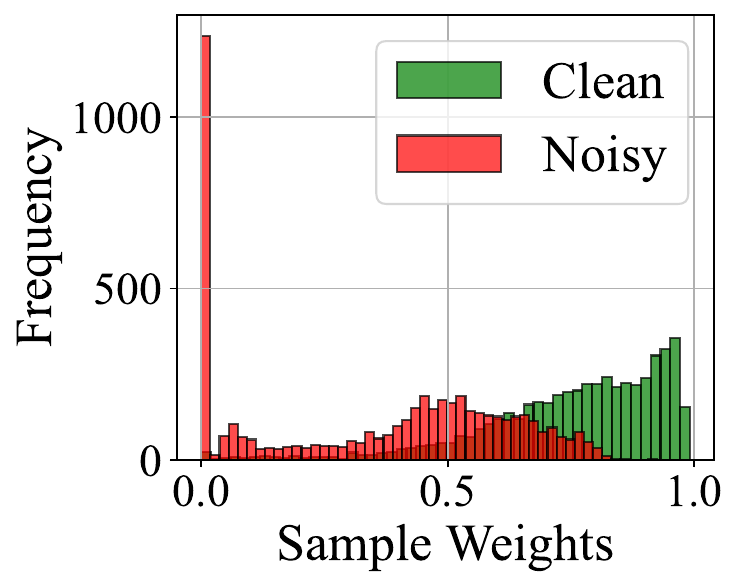}
        \caption{Initial distribution (epoch 5)}
    \end{subfigure}
    \hfill
    \begin{subfigure}[t]{0.48\linewidth}
        \centering
        \includegraphics[width=\linewidth]{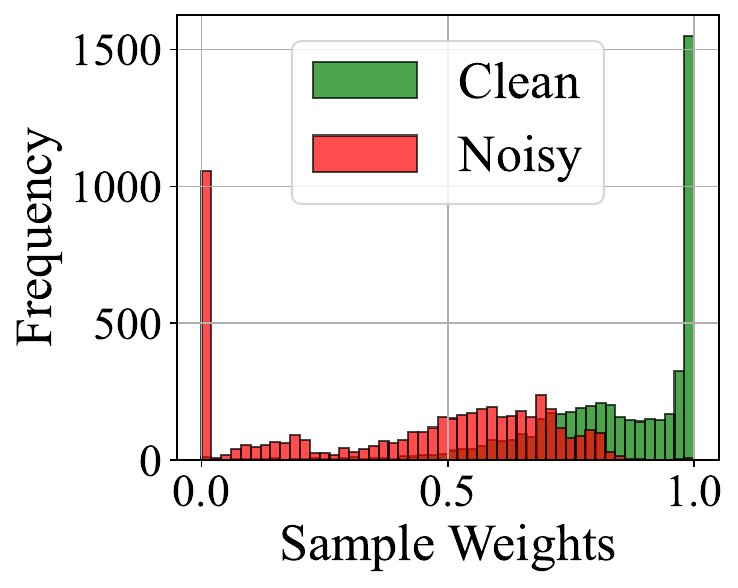}
        \caption{Final distribution (epoch 50)}
    \end{subfigure}
    
    \caption{Sample weight distributions for clean and noisy samples on the MS-COCO dataset with 64-bit hash codes under a 50\% noise rate.}
    \label{fig:weight_analysis}
\end{figure}

\subsubsection{Weight Strategy Analysis}

As illustrated in Figure \ref{fig:weight_analysis}, under 50\% noise rate on the MS-COCO dataset with 64 bits, the proposed weighting strategy effectively distinguishes clean from noisy samples. The transition from an initially uniform distribution to a bimodal pattern concentrated at both extremes confirms the precision of confidence assignment: reliable samples are assigned high weights, while potentially mislabeled instances are suppressed, thereby enhancing robustness against label noise.

\section{Conclusion}
In this paper, we propose a novel Semantic-Consistent Bidirectional Contrastive Hashing (SCBCH) framework to address the challenge of noisy supervision in multi-label cross-modal hashing.
SCBCH consists of two key modules: Cross-modal Semantic-Consistent Classification (CSCC) and Bidirectional Soft Contrastive Hashing (BSCH).
Specifically, CSCC leverages cross-modal semantic consistency to generate adaptive soft supervision, retaining both clean and noisy samples for discriminative learning while preserving potentially informative data. BSCH improves robustness by dynamically constructing balanced soft pairs based on label overlap, thereby making full use of fine-grained semantic correlations and reducing the adverse impact of noisy annotations.
Experiments on four benchmarks show SCBCH outperforms 12 state-of-the-art methods under noise, setting a new robust cross-modal hashing baseline.

\section{Acknowledgements}
This work was supported by the National Natural Science Foundation of China (62306197, 62372315), Sichuan Science and Technology Planning Project (2025ZNSFSC1507, 2024YFG0007, 2024ZDZX0004, 2024NSFTD0049), Central Government's Guide to Local Science and Technology Development Fund (2025ZYDF101), China Postdoctoral Science Foundation (2021TQ0223, 2022M712236), Chengdu Science and Technology Project (2023-XT00-00004-GX), Postdoctoral Joint Training Program of Sichuan University (SCDXLHPY2307).

\bibliography{main}

\end{document}